\documentclass[conference]{IEEEtran}
\IEEEoverridecommandlockouts
\usepackage{cite}
\usepackage{amsmath,amssymb,amsfonts}
\usepackage{algorithmic}
\usepackage{graphicx}
\usepackage{textcomp}
\usepackage{xcolor}

\usepackage[utf8]{inputenc}   
\usepackage{subcaption}
\usepackage{hyperref}       
\usepackage{url}            
\usepackage{booktabs}       

\def\BibTeX{{\rm B\kern-.05em{\sc i\kern-.025em b}\kern-.08em
    T\kern-.1667em\lower.7ex\hbox{E}\kern-.125emX}}
\begin{document}

\title{Prune and Replace NAS}

\author{
	\IEEEauthorblockN{Kevin Alexander Laube}
	\IEEEauthorblockA{\textit{Cognitive Systems Group}\\
		\textit{University of Tuebingen}\\
		Tuebingen, Germany\\
		kevin.laube@uni-tuebingen.de}
	\and
	\IEEEauthorblockN{Andreas Zell}
	\IEEEauthorblockA{\textit{Cognitive Systems Group}\\
		\textit{University of Tuebingen}\\
		Tuebingen, Germany\\
		andreas.zell@uni-tuebingen.de}
}

\maketitle

\begin{abstract}
	While recent Neural Architecture Search (NAS) algorithms are thousands of times faster than the pioneering works, it is often overlooked that they use fewer candidate operations, resulting in a significantly smaller search space.
	We present PR-DARTS, a NAS algorithm that discovers strong network configurations in a much larger search space and a single day. A small candidate operation pool is used, from which candidates are progressively pruned and replaced with better performing ones.
	Experiments on CIFAR-10 and CIFAR-100 achieve 2.51\% and 15.53\% test error, respectively, despite searching in a space where each cell has 150 times as many possible configurations than in the DARTS baseline.
	All of our code is available at \url{https://github.com/cogsys-tuebingen/prdarts}
\end{abstract}



\section{Introduction}
\label{s_introduction}

Since the groundbreaking results of AlexNet \cite{net_alex} in image classification, machine learning research has shifted from handcrafting features to handcrafting better network topologies. Architectures such as ResNet \cite{net_res}, DenseNet \cite{net_dense}, PyramidNet \cite{net_pyr} or MobileNetV2 \cite{net_mobv2} improved the performance on popular image datasets, at a fraction of the computational costs of earlier models.

Following the pioneering work of Zoph and Le on Neural Architecture Search \cite{nas}, the next shift is taking place.
Automatically designed networks, created from handcrafted search algorithms, have improved over their handcrafted competition in accuracy, FLOPs, and measured latency.
Unlike before, architectures can be automatically optimized for different metrics, datasets, target hardware, and under resource constraints, saving researchers countless hours of trial and error.
However, even though they contain billions of possible configurations, automatically designed architectures are limited by their respective search space definitions, such as a small set of candidate operations \cite{nas_trans, nas_evo, nas_prog, nas_enas, nas_darts, nas_sharp, nas_bench, nas_pdarts, nas_mde, nas_asap}.

We explore a much broader set of candidate operations, which is not initially purged of unsuccessful candidates from prior experiments \cite{nas_trans, nas_evo}.
The search is efficiently guided by progressively pruning bad candidates from a small pool, then replacing them with operations that arise from the better performing ones.
Network morphisms \cite{morph} enable us to change filter sizes, expansion ratios \cite{net_mobv2}, and the dilation of convolutions, while being able to use the learned weights of their respective parent operations.
We apply our Prune and Replace method to DARTS \cite{nas_darts}, thereby enlarging the search space by over 150 times per cell, and make only the necessary algorithmic adjustments for the search process to work. Despite the lack of prior operation selection, the resulting PR-DARTS algorithm is competitive with other recent NAS methods, furthering the efforts in fully automating architecture search.


\section{Background and related Work}
\label{s_related_work}

\subsection{Architecture search}
\label{ss_rel_nas}

The problem of architecture search is to find the network design that maximizes an objective function on the target task, such as accuracy or latency.
While initially prohibitively expensive \cite{nas, nas_trans, nas_evo, nas_prog}, recent algorithms find good architectures in GPU days or even hours \cite{nas_enas, nas_darts, nas_sharp, nas_mde, nas_asap, nas_pdarts}.

\subsubsection*{Micro search space}
\label{sss_rel_micro}

Instead of optimizing the architecture of the whole network (i.e. \textit{macro search space}), searching for a repeatable structure, named cell, has multiple advantages.
Firstly, the search space is significantly smaller, resulting in a problem that is easier to solve.
Furthermore, cells can be searched on smaller proxy networks and proxy datasets to speed up the search process, later being transferred to the target task \cite{nas_evo, nas_trans, nas_prog, nas_enas, nas_darts, nas_sharp, nas_mde, nas_asap, nas_pdarts}.
Typically two different cells are searched at the same time. A \textit{normal cell} that keeps spatial resolution and channel sizes the same, and a \textit{reduction cell} that halves the spatial and increases the channel size.

\subsubsection*{Weight sharing}
\label{sss_rel_weightsharing}

Training thousands of models \cite{nas, nas_trans, nas_evo, nas_prog} is expensive and inefficient, as most results are thrown away.
The concept of sharing weights across different architecture configurations has been introduced by ENAS \cite{nas_enas} and is now a core component of many recent NAS algorithms \cite{nas_enas, nas_darts, nas_sharp, nas_mde, nas_asap, nas_pdarts}.
These algorithms use a single over-complete model,
which contains all possible architecture configurations at the same time,
throughout the search process.
Different configurations can be trained and tested by using their corresponding subsets of the available weights. As the subsets of different configurations overlap, training a specific configuration also influences many others.

\subsubsection*{Search strategies}
\label{sss_rel_searchstrategy}

Even when employing weight sharing in a micro search space, most search spaces still contain billions of possible configurations. Proposed strategies to find promising configurations more efficiently include reinforcement learning \cite{nas, nas_trans, nas_enas}, evolutionary algorithms \cite{nas_evo}, gradient based optimization \cite{nas_darts, nas_asap, nas_sharp, nas_pdarts}, distribution learning \cite{nas_mde} and more.
Further improvements can be achieved by progressively increasing the model size \cite{nas_prog, nas_pdarts} or pruning bad candidate operations \cite{nas_asap}.

\subsubsection*{NAS properties}
\label{sss_rel_properties}

Active research on NAS unveils interesting properties that are relevant for our experiments:
\begin{enumerate}
	\item {
		Performance ranking hypothesis:
		''If Cell A has higher validation performance than Cell B on a specific network and a training epoch, Cell A tends to be better than Cell B on different networks after the training of these networks converge.'' \cite{nas_asap}	
	}
	\item {
		Locality:
		The performance of two different cells correlates with their edit distance, so that similar cells tend to have a similar performance. The effect vanishes at an edit distance of around 6. \cite{nas_bench}
	}
	\item {
		Depth Gap:
		Searching optimal cells in a shallow proxy network results in cells that may be suboptimal in deeper networks. \cite{nas_pdarts}
	}
\end{enumerate}

\subsection{DARTS}
\label{ss_rel_darts}

DARTS \cite{nas_darts}, the first gradient based NAS algorithm, uses an over-complete model to find two cells in the micro search space.
Each cell is a directed acyclic graph (DAG) of \textit{M} nodes $\{B_1, ..., B_M\}$, and takes inputs from 2 previous cells.
Each node $B_j$ is connected (has a directed edge) to all previous nodes and cell inputs of the same cell during the search phase, and computes the sum $x^{(j)}$ over its inputs.

\vspace*{-2pt}
$$x^{(j)} = \sum_{i < j} o^{(j, i)}(x^{(i)})$$

\vspace*{-4pt}
$$
\bar{o}^{(j, i)}(x) = \sum_{o \in O}
\frac{exp(\alpha_o^{(j, i)})}{\sum_{o' \in O} exp(\alpha_{o'}^{(j, i)})}
o(x)$$

\vspace*{0pt}
To include all candidate operations on each graph edge, DARTS relaxes the search space, computing the weighted sum $\bar{o}^{(j, i)}(x)$ over the outputs of all candidates $o \in O$. This is parametrized by the architecture weights $\alpha$, so that the cell search becomes differentiable and reduces to finding the weights $\alpha$ that maximize the model performance on the given task.
When finalized, each node $B_j$ takes exactly two inputs, and each $o^{(j, i)}$ consists of a single operation. To achieve this, DARTS simply uses the highest weighted options at the end of training.
The cell output is the average of all nodes.

As there are two active graph edges per node, each with K candidates, 
the total number of possible configurations per cell is $\prod_{n=2}^{M+1} {n \choose 2} \cdot K^2 \approx 9.3 \cdot 10^{9}$, 
with the standard search parameters ($M=4$ and $K = \vert O \vert =8$). While the gradient based optimization enables a comparably quick and stable search process, calculating the outputs of all candidate operations requires considerable amounts of FLOPs and memory.

\subsection{Network morphisms}
\label{ss_rel_morphisms}

For neural networks, a morphism is a function preserving operation that transfers knowledge from one network into a new, generally more powerful network \cite{morph, morph_net2net}.
Standard operations are increasing the kernel size of convolutions or widening a layer, or even inserting new computation paths in parallel, which are summed or concatenated with existing graph nodes \cite{morph_pathlevel}.
The common applications for network morphisms are transfer learning \cite{morph_net2net} and fine-tuning architectures \cite{morph}, but they have also been applied in NAS \cite{morph_nas, morph_pathlevel}.


\section{Methods}
\label{s_methods}

We provide an overview in Sect. \ref{ss_methods_searchspace}, explain how we explore vast operation spaces in Sect. \ref{ss_methods_prog} and finally how we cope with varying hardware requirements in Sect. \ref{ss_methods_hardware}.

In a nutshell, we perform several short cell optimization iterations, called cycles, which are comparable to a DARTS search. In each cycle we prune bad candidate operations and replace them with more promising ones. This bears similarity to P-DARTS \cite{nas_pdarts} where operations are pruned in progressively larger networks, or other algorithms that prune operations in a single cycle \cite{nas_asap}. To the best of our knowledge, we are however the first ones to insert new candidate operations during the search process.

\subsection{Search Overview}
\label{ss_methods_searchspace}

Our algorithm uses the classic DARTS search space, except for the available candidate operations.
We search for a normal and a reduction cell, each cell takes the two prior ones as inputs, has $M=4$ nodes, the cell output being their mean.
As in DARTS, we use gradient descent to optimize the network and architecture weights.

We train the network and architecture weights for a certain number of epochs in each optimization cycle. Since we are mostly interested in the ranking of the candidates, not their optimal probabilities, a small number of epochs per cycle suffices.
To avoid punishing operations that are harder to learn, we begin each cycle with grace epochs, in which the architecture weights are frozen. After training we remove a number of bad candidate operations and replace them with morphisms of better performing ones. This is done on a per-edge case on the cell graph, so that the number of different candidate operations of the whole cell graph can be significantly larger than that of any single edge.
As we prune more candidate operations than we add in the later cycles, the cells converge to their finalized configurations. One specific search process from our experiments is displayed in Figure \ref{fig_training}.

\subsection{Exploring vast operation spaces with morphisms}
\label{ss_methods_prog}
\label{sss_methods_morphing}

\begin{figure}[htbp]
	\centerline{\includegraphics[trim=96 620 240 94, clip, scale=0.8]{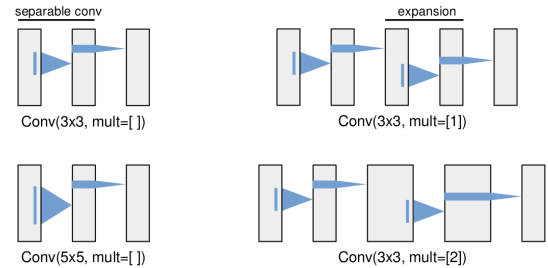}}
	\caption{
		An example how our initial separable convolution candidate operation (top left) can be morphed by increasing the kernel size (bottom left) or inserting an inverted bottleneck expansion \cite{net_mobv2} (top right). Finally, a second morphism can widen an existing expansion (bottom right). \textit{mult} describes the width of the expansion ratios, as a sequence of channel-multiples with respect to the initial layer's channel count.
	}
	\label{fig_conv}
\end{figure}

We consider the following operations on separable convolution layers that do not change the output:
\begin{enumerate}
	\item[1.] {
		Increasing the kernel size by padding it with zeros, see Fig. \ref{fig_conv} (bottom left).
	}
	\item[2.] {
		Widening a layer, as depicted in Fig. \ref{fig_conv} (bottom right). The weights using the new neurons are initialized with zeros. This can only be done when the convolution has at least one expansion layer, and always increases the currently smallest expansion layer factor by 1, preferring the leading one when multiple are equally wide. This expansion factor defines the width of a layer, as a multiple of the candidate's initial channel count. 
	}
\end{enumerate}
And further operations that, as we use them, are strictly speaking not network morphisms:
\begin{enumerate}
	\item[3.] {
		Inserting layers with non-linearities, as depicted in Fig. \ref{fig_conv} top right. Although parametrized non-linear functions can keep the identity operation \cite{morph} we refrain from using them. New expansion layers are always inserted at the back with width factor 1.
	}
	\item[4.] {
		Decreasing the kernel size by using only their spatial center weights.
	}
	\item[5.] {
		Increasing the dilation, thus widening the kernel's receptive field by using spatially distant inputs.
	}
	\item[6.] {
		Decreasing the dilation.
	}
\end{enumerate}
As we impose restrictions on the search space, such as a maximum or minimum kernel size, the number of available morphisms per operation is usually lower. All kernels of one candidate operation use the same size and dilation value.

When a candidate operation is selected as parent, one of the available morphisms is picked uniformly at random to create a morphed child.
If the resulting candidate operation is or was already part of the candidate pool of the respective cell graph edge, a new parent operation will be sampled.
As some outcomes are not true morphisms but need to be adapted, we continue the search process with \textit{grace epochs}, during which only network weights are trained.

\subsubsection{The group similarity problem}
\label{sss_methods_groupproblem}

While using network morphisms enables us to reuse trained weights in almost every cycle, we also have to consider the inherent disadvantage that we call \textit{group similarity problem}. After inserting morphed candidate operations and continuing training, we can expect the morphed candidates and their parents to be similar, especially since we reuse the parent's weights.
As this group of candidates has similar outputs, the DARTS algorithm can be expected to assign lower weights to them individually, as each operation is somewhat redundant. We consider two implications:

Firstly, pruning candidates has to be done more carefully. Even if all candidates $A_i$ from a group are superior to some other candidate $B$, due to the similarity problem, $B$ may have a higher architecture weight and may therefore wrongly be considered better.
This means that, if we remove more candidates than we inserted in the previous cycle, we may outright delete an entire promising group of likely candidates. 

Secondly, if the selection of parents for new morphed candidates is based on architecture weights, it is also affected. Especially greedy algorithms may only take the best performing candidate after each cycle, which is likely not representative.

\subsubsection{Iteratively pruning and replacing operations}
\label{sss_methods_rep}

We follow DARTS and initialize a small search network with a set of candidate operations on every graph edge. At the end of every cycle, we perform the following steps for every graph edge in both cells:

We first remove the worst performing candidate operations from this edge's candidate pool. The number is chosen so that, after the following insertion step, the desired amount of different candidates is available.
Note that, except when finalizing the architecture, we guarantee at least one convolution operation to remain in the candidate pool.
We then sample parents from the remaining candidates, from which morphed versions will be created to extend our pool. To do so, we randomly pick candidates distributed by the Softmax function over the architecture weights.
As at least one convolution remains in the pool, we are always guaranteed to find a candidate that can be morphed. However, when the proposed child is or was already part of the pool, we only accept it when further morphing attempts find no yet undiscovered candidate operation configurations.
Thus, every edge develops its own pool of candidates over time.

In the next cycle, we load available weights for every remaining candidate and initialize the new ones from their respective parents. The architecture weights are reset, so that all candidates are initially equally weighted.

\subsection{Changing hardware requirements}
\label{ss_methods_hardware}

As the available candidate operations change over time, with an expected shift towards using more expensive ones during the search process, we will need more and more memory. 
While we could initialize the search process in a way that prevents out-of-memory problems in the later search stages, doing so from the start would be wasteful and possibly slow.

Instead, we adopt a simple pragmatic approach of scaling the batch size depending on available GPU memory. Given a minimum batch size $b_{min}$ and a value $b_{mult}$, we try to find $n$ that maximizes the batch size $b_{min} + n \cdot b_{mult}$, limited by the maximum batch size $b_{max}$, until the expected memory consumption surpasses a set threshold, e.g. 95\%.
We find that assuming a linear relationship between batch size and required GPU memory works reasonably well and is simple to implement. As this process is very cheap, we try to increase the batch size every five training steps, and reset the batch size to the minimum value when we replace candidate operations.


\section{Experiments}
\label{s_experiments}

We detail the search and training configuration for our experiments in Sect. \ref{ss_experiments_details}, analyze one search process in Sect. \ref{ss_results_search} and finally list the retraining results in Sect. \ref{ss_results_retraining}.
All of our experiments were run on a single Nvidia GTX 1080 Ti GPU.

\begin{table}[htbp]
	\caption{Search parameters for each cycle, see Fig. \ref{fig_training} for a visual representation.}
	
	\centering
	\begin{center}
		\begin{tabular}{ l  c c c  c c c c c c }
			\toprule
			& \multicolumn{9}{c}{\textbf{cycle}} \\
			\cmidrule(r){2-10}
			\textbf{\textbf{parameter}} & \textbf{1} & \textbf{2} & \textbf{3} & \textbf{4} & \textbf{5} & \textbf{6} & \textbf{7} & \textbf{8} & \textbf{9} \\
			\midrule
			epochs 			& 15 & 15 & 10 & 10 & 10 & 10 & 10 & 10 & 10 \\
			grace epochs 	& 5  & 5  & 3  & 3  & 3  & 3  & 3  & 3  & 3  \\
			morphisms		& 3  & 3  & 3  & 3  & 3  & 0  & 0  & 0  & 0  \\
			candidates		& 6  & 6  & 6  & 6  & 6  & 4  & 3  & 2  & 1  \\
			\bottomrule
		\end{tabular}
		\label{tab_search}
	\end{center}
\end{table}

\subsection{Details}
\label{ss_experiments_details}

\subsubsection*{General search settings}

We follow the DARTS setup as far as possible.
The cells are searched in a proxy network of eight cells, six normal cells with a reduction cell inserted at the first and second thirds of the network, the first cell has a channel count of 16.
We use stochastic gradient descent (SGD) with learning rate $0.025$, which is annealed via cosine decay to a minimum of $0.01$ at the end of each cycle, a momentum term of $0.9$ for the network weights, and a weight decay of $0.0003$.
The architecture weights are trained using Adam \cite{etc_adam} with a constant learning rate of $0.0006$, $\beta _1 = 0.5$, $\beta_2 = 0.999$, and a weight decay of 0.001.
We use CIFAR-10 \cite{etc_cifar} to search the cells, a popular image dataset of $50,000$  training and $10,000$ test images. Each image belongs to exactly one of the ten classes and has a size of $32 \times 32$ pixels in RGB colors. The training set is split in half, using one half to train the network weights and the other half for the architecture weights. Images are normalized by the mean and standard deviation of the dataset, randomly horizontally flipped, zero padded to $40 \times 40$ and randomly cropped back to $32 \times 32$.

Unlike DARTS we use a minimum learning rate, warm restarts over multiple cycles, and a variable batch size. The parameters that change per cycle are detailed in Table \ref{tab_search} and depicted in Figure \ref{fig_training}. The design scheme is simple and not optimized.
We split a total of 100 epochs into different cycles, whereas the first two cycles are slightly longer, since the weights are not trained and the most relevant pruning actions take place, and use a third of each cycle as a grace period.
We replace half of the available candidates with new ones until including cycle five, after which we only prune until convergence.
Furthermore, our initial configuration space contains only four operations: $3 \times 3$ max and average poolings, the identity function (factorized reduction when the stride is 2) and a $3 \times 3$ separable convolution that is, unlike as in DARTS or other NAS algorithms, not stacked twice.

\subsubsection*{Restrictions for morphisms}

To test our algorithm in different search spaces, we add constraints to how a candidate convolution operation may be morphed, and list them in Table \ref{tab_constraints}. While pooling operations could also vary e.g. in their kernel sizes, we excluded that direction in our current research for simplicity.
Note that our \textit{DARTS-like} space contains more configurations than the original, since we use all combinations of kernel size, dilations and expansions of depth 0 or 1 (convolution not stacked or stacked once).
We then also lift the width restriction in the \textit{restrict depth} space and add a $1 \times 1$ kernel.
Finally, the \textit{unrestricted} space has no restrictions with respect to expansion depth or width.

\begin{table*}[htbp]
	\caption{Morphing constraints for our experiment settings, limited to convolutions. The total number of configurations includes only the obtainable configurations as described in Sect. \ref{sss_methods_morphing} within five morphing steps, and excludes all variations of kernel size $k=1$ combined with dilation $d=2$.}
	\centering
	\begin{center}
		\begin{tabular}{ l  c c  c c  c c  c }
			\toprule
			& \multicolumn{2}{c}{\textbf{kernel sizes}} & \multicolumn{2}{c}{\textbf{dilations}} & 
			\multicolumn{2}{c}{\textbf{expansions}} & \\
			\cmidrule(r){2-3}
			\cmidrule(r){4-5}
			\cmidrule(r){6-7}
			\textbf{\textbf{setting}} & \textbf{min} & \textbf{max} & \textbf{min} & \textbf{max} & \textbf{max depth} & \textbf{max width} & \textbf{total} \\
			\midrule
			DARTS-like (DL)			& 3  & 7  & 1  & 2  & 1  & 1 & 12 \\
			depth-restricted (DR)	& 1  & 7  & 1  & 2  & 1  & 5 & 36 \\
			unrestricted (UR)		& 1  & 7  & 1  & 2  & -  & - & 80 \\
			\bottomrule
		\end{tabular}
		\label{tab_constraints}
	\end{center}
\end{table*}

\subsubsection*{Retraining}

The process of retraining the evaluation architecture is identical to DARTS. A model of 36 initial channels and 20 cells is built, 18 normal cells with reduction cells inserted at the first and second third. The model is trained on the full training set and later evaluated on the test set.
An auxiliary head is inserted at two thirds of the model and weighted with $0.4$. The model is trained using SGD of learning rate $0.025$ which is annealed to $0$ using cosine decay over 600 epochs, momentum of $0.9$ and weight decay of $0.0003$.
Drop-path \cite{etc_droppath} of linearly increasing probability up to $0.3$, and Cutout \cite{etc_cutout} using $16 \times 16$ pixel squares are used in addition to the previous regularization methods.

\begin{figure*}[htbp]
	\centerline{\includegraphics[trim=10 45 10 70, clip, scale=0.57]{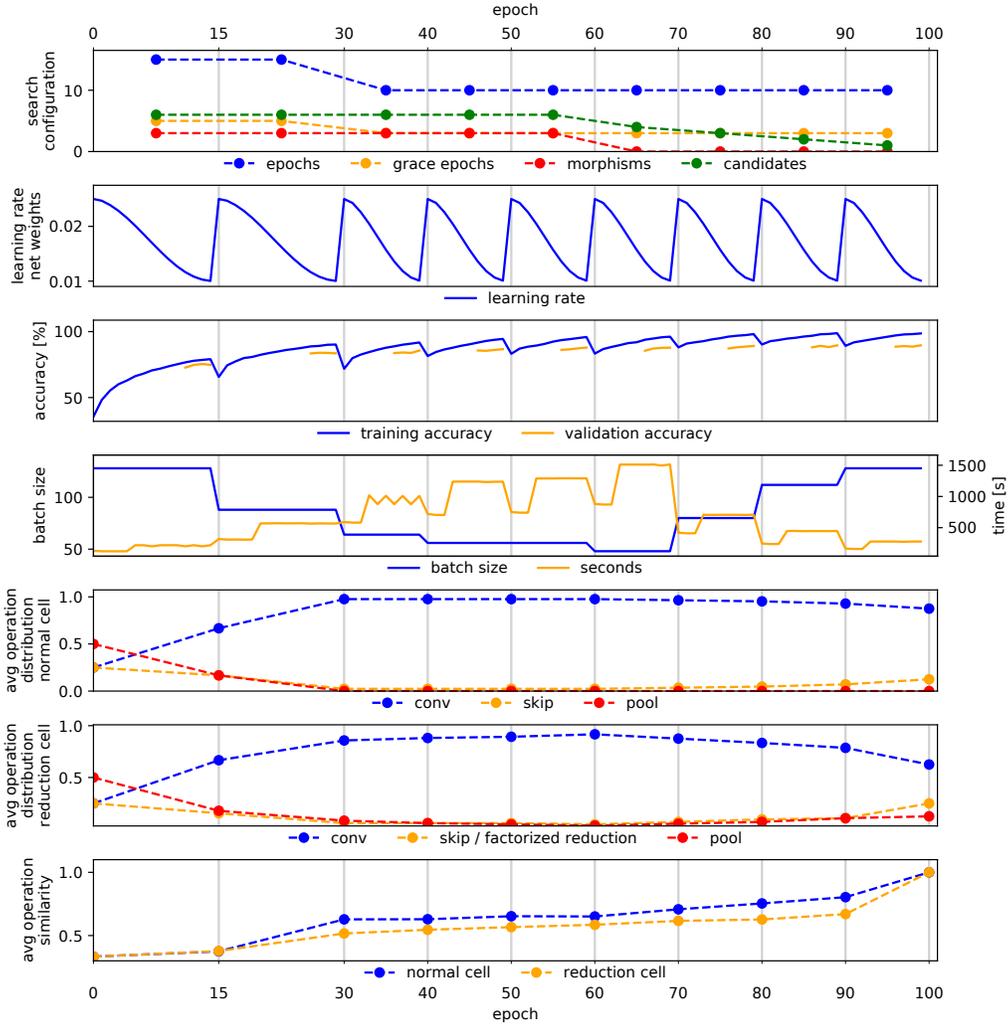}}
	\caption{
		The search process of our PR-DARTS DL2 cells, best viewed in color.
	}
	\label{fig_training}
\end{figure*}

\subsection{Search results}
\label{ss_results_search}

We visualize the search process of the DL2 model in Figure \ref{fig_training} and include statistics that provide us with additional understanding:

\subsubsection*{Accuracy}
As to be expected, the training accuracy has minor drops after candidate operations are pruned and replaced, the architecture weights are reset, and the learning rate is increased again. The validation accuracy is slightly lower and converges to around 90\%.

\subsubsection*{Batch size and time}
As cheaper candidates (e.g. identity or pooling) are pruned and replaced with more costly convolutions, the required memory increases. This is compensated by a smaller batch size, which decreases to as little as 48, at the cost of increasingly longer epoch durations. As the number of candidates decreases, the batch size can be increased again.

\subsubsection*{Distribution of operations}
\begin{figure*}[htbp]

	\begin{subfigure}{\textwidth}
		\begin{subfigure}{.48\textwidth}
			\includegraphics[trim=0 0 0 0,clip,width=\textwidth]{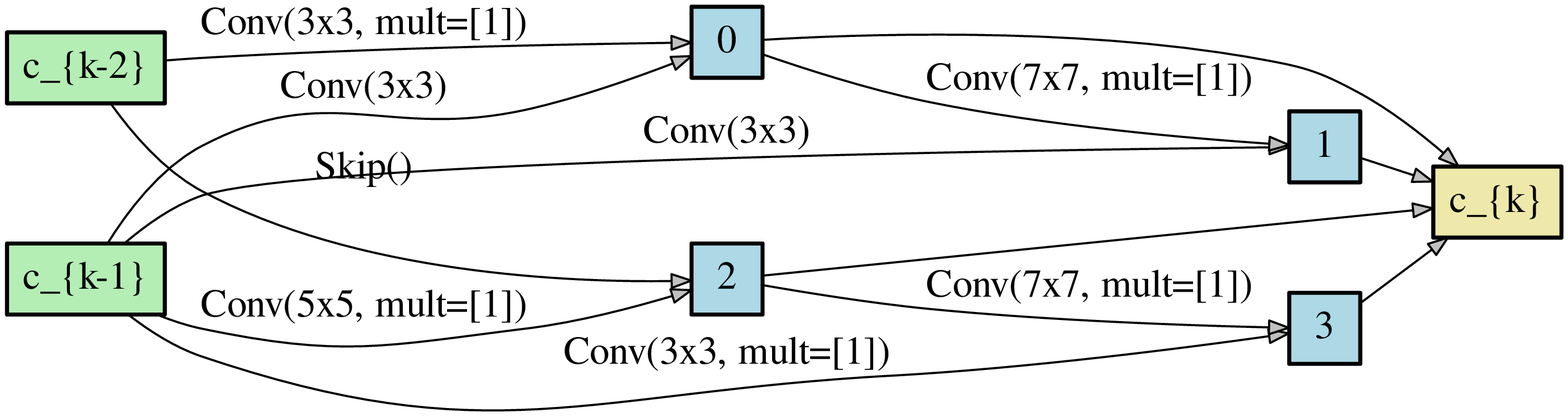}
		\end{subfigure}
		\hfill
		\begin{subfigure}{.48\textwidth}
			\includegraphics[trim=0 0 0 0,clip,width=\textwidth]{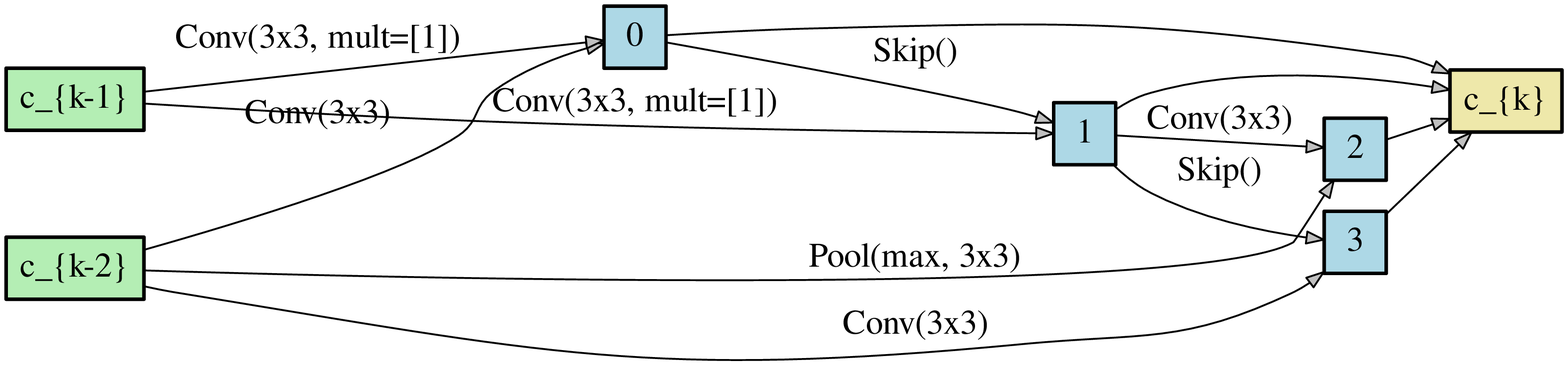}
		\end{subfigure}
		\caption{Normal and reduction cell in the DARTS-like setting DL2.}
		\label{fig_cells_dl2}
	\end{subfigure}
	
	\begin{subfigure}{\textwidth}
		\begin{subfigure}{.48\textwidth}
			\includegraphics[trim=0 0 0 0,clip,width=\textwidth]{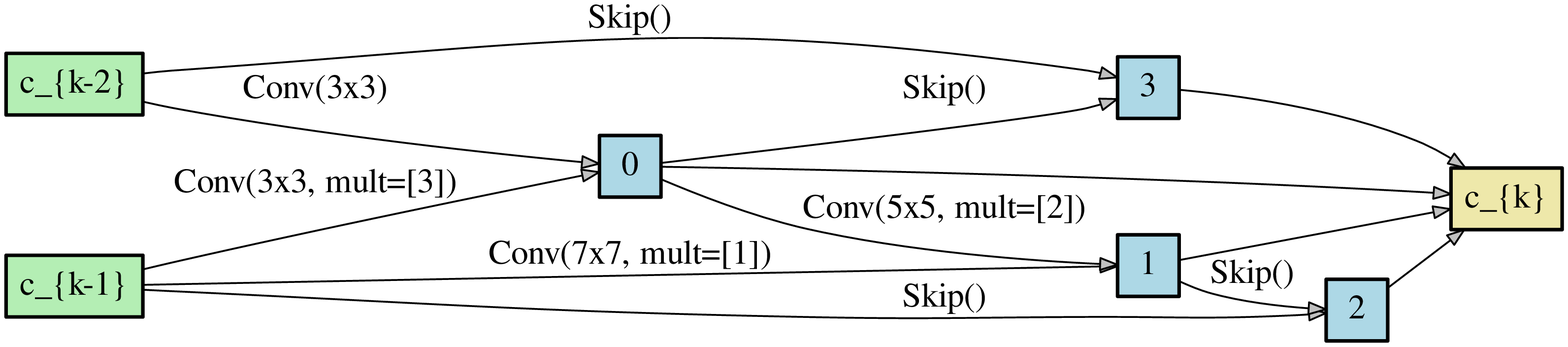}
		\end{subfigure}
		\hfill
		\begin{subfigure}{.48\textwidth}
			\includegraphics[trim=0 0 0 0,clip,width=\textwidth]{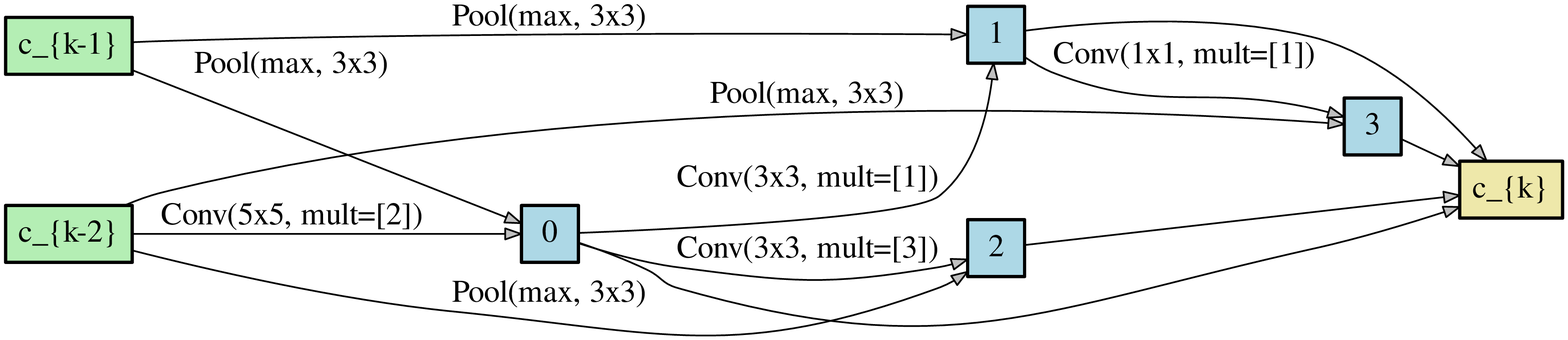}
		\end{subfigure}%
		\caption{Normal and reduction cell in the depth-restricted setting DR.}
		\label{fig_cells_rd}
	\end{subfigure}

	\caption{Discovered cells in the presented experiments in Table \ref{tab_results}, all convolutions are separable. The nodes $B_j$ (blue) sum their inputs, the average of all nodes is the cell output (orange). The operations are plotted above their respective graph edges.}
	\label{fig_cells}
\end{figure*}

Starting with two poolings, one identity and one convolution candidate, the initial pooling ratio is $50\%$.
As operations are pruned and added, the ratios of skip and poolings drops. When the number of candidates is gradually reduced, their ratio slowly increases again.
As depicted in Figures \ref{fig_training} and \ref{fig_cells_dl2}, a single identity candidate in the normal cell survived every pruning step.

\subsubsection*{Operation similarity}
We introduce a simple measure of \textit{similarity} between two candidates: If they belong to different types (id, pool, conv), the \textit{similarity} is zero. It is otherwise set to $(1 + k) / (1 + n)$ for $k$ out of $n$ equal configuration parameters (pool type, kernel size, dilation), expansion ratios may be partly similar.
The \textit{similarity} over $m$ candidates is the average over all pairwise \textit{similarity} values.

We observe that the \textit{similarity} value is rising for both cells, even when the ratio of convolution candidates drops.
As the small number of remaining identity and pooling candidates is limited to few cell graph edges and reduce average \textit{similarity}, the candidate pools on other edges must gradually lose diversity. This hints that we could possibly converge to the final architecture in fewer pruning cycles or that the group similarity problem (see Sect. \ref{sss_methods_groupproblem}) has less influence than expected.

\begin{table*}[htbp]
	\caption{Test error on CIFAR-10 and CIFAR-100, all results are using standard regularization (flipping, shifting, weight decay), drop-path \cite{etc_droppath}, and Cutout \cite{etc_cutout}. Methods that use other additional regularization techniques are excluded. We list the numbers of different discovered$^\dagger$ and discoverable$^\ddagger$ operations of our normal cells, of which some are visualized in Figure \ref{fig_cells}.}
	
	\centering
	\begin{center}
		\begin{tabular}{ l  c c c  c c c }
			\toprule
			& & \multicolumn{2}{c}{\textbf{CIFAR test error [\%]}} & \multicolumn{3}{c}{\textbf{Search}} \\
			
			\cmidrule(r){3-4}
			\cmidrule(r){5-7}
			\textbf{Method} & \textbf{\#params} & \textbf{10} & \textbf{100} & \textbf{GPU days} & \textbf{\#ops} & \textbf{method} \\
			\midrule
			NASNet-A~\cite{nas_trans} 			& 3.3M & $2.65$	& $ $	  & 1800 & 13 & RL \\
			AmoebaNet-B~\cite{nas_evo} 			& 2.8M & $2.55$ & $ $	  & 3150 & 19 & evo\\
			ENAS~\cite{nas_enas}				& 4.6M & $2.89$	& $ $	  & 0.5  & 5  & RL \\
			\midrule
			DARTS (1st order)~\cite{nas_darts} 	& 2.9M & $2.94$	& $ $	  & 1.5  & 8  & grad \\
			DARTS (2nd order)~\cite{nas_darts} 	& 3.4M & $2.83$ & $ $	  & 4.0  & 8  & grad \\
			P-DARTS C10~\cite{nas_pdarts} 		& 3.4M & $2.50$	& $16.55$ & 0.3  & 8  & grad \\
			P-DARTS C100~\cite{nas_pdarts} 		& 3.6M & $2.62$	& $15.92$ & 0.3  & 8  & grad \\
			MDENAS~\cite{nas_mde} 				& 4.1M & $2.40$	& $ $	  & 0.16 & 8  & MDE \\
			sharpDARTS~\cite{nas_sharp} 		& 3.6M & $2.45$	& $ $	  & 0.8  &    & grad \\
			SNAS moderate~\cite{nas_snas}	 	& 2.8M & $2.85$	& $ $	  & 1.5  & 8  & grad \\
			NASP~\cite{nas_proximal} 			& 3.3M & $2.80$	& $ $	  & 0.2  & 7  & grad \\
			NASP (more ops)~\cite{nas_proximal}	& 7.4M & $2.50$	& $ $	  & 0.3  & 12 & grad \\
			
			\midrule
			PR-DARTS DL1					& 3.2M & $2.74 \pm 0.12$ & $17.37 \pm 0.14$  & 0.82 & 15$^\dagger$/15$^\ddagger$ & grad \\
			PR-DARTS DL2					& 4.0M & $2.51 \pm 0.09$ & $15.53 \pm 0.29$  & 0.82 & 15$^\dagger$/15$^\ddagger$ & grad \\
			PR-DARTS DR					& 4.2M & $2.55 \pm 0.19$ & $16.69 \pm 0.08$  & 0.88 & 26$^\dagger$/39$^\ddagger$ & grad \\
			PR-DARTS UR					& 5.4M & $3.79 \pm 0.24$ &  & 1.10 & 45$^\dagger$/83$^\ddagger$ & grad \\
			\bottomrule
		\end{tabular}
		\label{tab_results}
	\end{center}
\end{table*}

\subsection{Retraining results}
\label{ss_results_retraining}

The retraining results are listed in Table \ref{tab_results} and their corresponding cells depicted in Figure \ref{fig_cells}. We report the average test accuracy over the last five epochs, averaged over three independent runs with different seeds. The reported results stem from the most interesting experiments out of three independent search process, per restriction configuration.

Despite increasing the number of candidate operations, thus increasing the difficulty of the task, our DL1 model surpasses its baseline and is competitive with other methods that improve over it. Interestingly, all operations in the DL1 model are also obtainable by regular DARTS. We hypothesize the progressive pruning to contribute most to the improvement.
DL2 further increases the accuracy, especially on CIFAR-100, at the cost of more parameters. Even with considerably more candidates, the DR model performs nearly similar to DL2 on CIFAR-10 and adopts some newly available candidates in the final cells, such as a convolution with a $1 \times 1$ kernel and expansion ratios greater than $1$. 

However the unrestricted model UR suffers from a significant performance drop in accuracy and the number of parameters. The small search model benefits from multiple operations are deeply stacked, which is a suboptimal choice for the 2.5 times larger evaluation model. This depth gap \cite{nas_pdarts} has been observed before and alleviated by increasing the search model size throughout the process, as the number of candidates is reduced. As we only use a small model, with a  much larger search space, the decrease in performance is hardly surprising.

We note that none of the discovered cells uses a convolution dilation greater than 1. While such operations are possibly unsuitable for the CIFAR-10 dataset, reusing weights when changing the dilation factor may require additional care.


\section{Discussion and Future Work}
\label{s_discussion}

While the first NAS algorithms experimented with a huge number of candidate operations \cite{nas, nas_evo, nas_trans}, later algorithms \cite{nas_enas, nas_darts} restricted the search space to exactly those candidates that have already proven their benefits.
This design space change significantly simplifies the problem, thus speeds up search, but also implicitly improves the final evaluation performance \cite{etc_designspaces}.
Rather than further improving the results in this setting, we present the Prune and Replace method, applied to DARTS, to search through a much larger space in reasonable time and with comparable performance.
While there are roughly
$9.3 \cdot 10^{9}$
possible configurations in the DARTS search space per cell, our DARTS-like space is roughly 153 times larger with
$\prod_{n=2}^{5} {n \choose 2} \cdot 15^2 \approx 4.6 \cdot 10^{11}$ possible cell configurations. Naturally, the less restricted spaces are also much larger.

The presented PR-DARTS algorithm does not perform well when the depth is not restricted. This depth gap can be alleviated by using (progressively) larger models \cite{nas_pdarts} in the search process, or by applying our Prune and Replace method to an algorithm that does not require a proxy search network.

Given that PR-DARTS is the first algorithm that inserts candidate operations during the search process, we see further room for improvement.
As new candidates are randomly morphed from others, and on each graph cell independently, we expect to discover bad performing ones multiple times. Weighting the randomly chosen morphisms by their performance of the whole cell is one simple way to direct the candidate search. 
Furthermore, while we included several cycles that only prune the network to counter the expected group similarity problem, we abstained from including further improvements over the DARTS baseline.
We expect search speed and cell quality to improve by e.g. using a concrete distribution over the softmax \cite{nas_asap, nas_snas, etc_concrete}, or using MDE optimization over gradients \cite{nas_mde}.
Regularizing the model complexity e.g. by considering FLOPs or limiting the amount of available skip connections \cite{nas_pdarts} may improve the algorithm further. 
Finally, the number of cycles and their values for epochs, grace epochs, number of candidate operations and morphisms have been chosen intuitively and are not optimized.


\section{Conclusion}
\label{s_conclusion}

We presented the Prune and Replace method, applied as PR-DARTS, which is the first NAS algorithm that prunes bad candidate operations and replaces them with better ones. Experiments on CIFAR-10 and CIFAR-100 show that PR-DARTS finds well performing models in a significantly larger search space, with only the necessary algorithm changes compared to the DARTS baseline.
A variable batch size during search works empirically well and enables us to change the network and candidate operations, without further considering hardware implications.


\section*{Acknowledgments}

We would like to thank Maximus Mutschler, Hauke Neitzel and Jonas Tebbe for valuable discussions and feedback.

\end{document}